\documentclass[10pt,twocolumn,letterpaper]{article}

\usepackage{iccv}
\usepackage{times}
\usepackage{epsfig}
\usepackage{graphicx}
\usepackage{amsmath}
\usepackage{amssymb}
\usepackage{makecell}
\usepackage{multirow}

\usepackage[pagebackref=true,breaklinks=true,letterpaper=true,colorlinks,bookmarks=false]{hyperref}

 \iccvfinalcopy 

\ificcvfinal\fi

\begin{document}

\title{Knowledge Guided Bidirectional Attention Network for Human-Object Interaction Detection}

\author{
  Jingjia Huang\thanks{Equal contribution.} \thanks{Corresponding author.} \ \ \ \ \ \ \ \ \ \ \ \ \ \ \ \ \ Baixiang Yang$^{*}$
  \\
  School of Electronic and Computer Engineering\\
  Peking University\\
  \{jjhuang, bxyang\}@pku.edu.cn
}

\maketitle
\ificcvfinal\thispagestyle{empty}\fi

\begin{abstract}
   Human Object Interaction (HOI) detection is a challenging task that requires to distinguish the interaction between a human-object pair. Attention based relation parsing is a popular and effective strategy utilized in HOI. However, current methods execute relation parsing in a ``bottom-up" manner. We argue that the independent use of the bottom-up parsing strategy in HOI is counter-intuitive and could lead to the diffusion of attention. Therefore, we introduce a novel knowledge-guided top-down attention into HOI, and propose to model the relation parsing as a ``look and search" process: execute scene-context modeling (\ie look), and then, given the knowledge of the target pair, search visual clues for the discrimination of the interaction between the pair. We implement the process via unifying the bottom-up and top-down attention in a single encoder-decoder based model. The experimental results show that our model achieves competitive performance on the V-COCO and HICO-DET datasets.


\end{abstract}

\section{Introduction}

\begin{figure}[t]
\begin{center}
\includegraphics[width=0.8\linewidth]{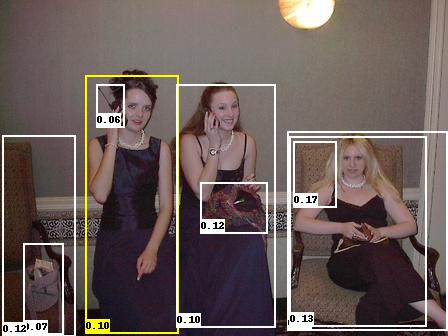}
\end{center}
   \caption{Visualization of the bottom-up attention distribution conditioned on the instance in the yellow box. The attention scores are attached to the bottom-left of the boxes. The model tends to pay equal attention to each instance in the scene, which illustrates the diffusion of the attention on the image.}
\label{motivation}
\end{figure}
In recent years, the computer vision community has paid increasing attention to understand real-word scenes from images, which requires not only answering the question ``What is where?" (\ie object recognition and localization), but also answering the question ``What is happening?" \cite{gao2018ican}. Human-Object Interaction (HOI) detection is a fundamental problem in the scene understanding. Given an image, HOI detection aims to identify all the triplets of the form $\langle human, verb, object\rangle$ in the image.

Relation parsing is widely adopted to improve the representation learning for the interaction recognition, where the relation is typically quantified by some forms of attention mechanisms (\eg self-attention \cite{vaswani2017attention}, graph structural parsing \cite{qi2018learning, liu2020consnet}). The representation of objects and humans that enhanced by the relation parsing will be then paired for the interaction recognition.

In current methods, most of the relation parsing is executed in a bottom-up manner. In the human visual system, the term ``bottom-up" refers to the way in which attention is focused automatically on signals associated with novel or salient stimuli, while ``top-down" means the manner in which the attention is focused volitionally on signals determined by specific tasks (\eg looking for evidences) \cite{buschman2007top,corbetta2002control}. In this paper, we adopt similar terminology and refer to the attention conditioned on each instance independently before pairing as ``bottom-up", which can be typically formulated as $P(A_{i}|S,i)$, where $i, A_{i}, S$ denote a detected instance (human or object), attention distribution conditioned on the instance and the scene, respectively.
Given the pairs that composed of the same person and different objects (\eg human typing on the computer while sitting on the chair), the attention of the model conditioned on the same human should be focused on different visual clues related to ``sit on chair" and ``type on computer", respectively. However, in the bottom-up attention, the attention distributions conditioned on the human for the two different pairs are identical, which is counter-intuitive.
It reveals that the bottom-up attention fails to collect customized relational information for the pairs directly. Moreover, it could lead to the diffusion of attention conditioned on the instance, especially in a complex scene. For example, as shown in Fig.\ref{motivation}, we employ self-attention for the relation parsing, which is one of the typical bottom-up attention mechanisms. We visualize the attention distribution conditioned on the woman (in yellow box), and it shows that the attention of the model fails to focus on the most related instances (\eg the phone) in the scene.

In order to handle this defect, we introduce the knowledge guided top-down attention into HOI, which has the flexibility to collect different information in the scene conditioned on given human-object pair $\langle h,o\rangle$. The knowledge refers to the information about the target human-object pair and possible interactions. The top-down attention can be formulated as $P(A_{\langle h,o\rangle, verb}|S,\langle h,o\rangle, verb)$, where $verb$ denotes the verb prior knowledge. We unify the top-down attention and bottom-up attention in a Knowledge Guided Bidirectional Attention Network (K-BAN).

 K-BAN models the relation parsing as a ``look and search" process: execute scene-context modeling (\ie look), and then, given the knowledge of the target pair, search visual clues for the discrimination of the interaction between the pair. K-BAN is constructed upon an encoder-decoder architecture. The bottom-up relation encoder serves for the scene context modeling, while the top-down decoder serves for visual clues searching. We introduce the human-object interactiveness knowledge into the context modeling via constructing the encoder with a newly proposed Group-aware Parsing Module (GPM). 
In the decoder, the detected humans and objects are paired as queries to selectively attend to the outputs of the encoder for clue mining. Each query is constructed with the semantic embedding of the object as well as the spatial relation coding of the corresponding object and human. Moreover, given an object, considering that different categories of possible interactions corresponding to the object can be characterized by different visual cues, a preferred decoder should have the capability to adapt its attention for the discrimination of different interactions.
Therefore, we further introduce the verb-object co-occurrence as the prior knowledge to guide the attention of the model through duplicating each query and augmenting them with the corresponding verb embeddings.
Extensive experiments are conducted on V-COCO and HICO-DET datasets, the results demonstrate the effectiveness of our method.

The main contributions of this work can be summarized as follows:
\begin{itemize}
\item To our best knowledge, we are the first to introduce the knowledge-guided top-down attention into the relation parsing for HOI detection.
\item Combining the knowledge-guided top-down relation parsing with a new group-aware bottom-up parsing strategy, we propose a novel model called K-BAN to improve the pair-wise representation learning in HOI.
\item K-BAN is conceptually simple and is able to achieve competitive performance on V-COCO and HICO-DET datasets.
\end{itemize}
\section{Related Work}
Our work is related to the work \cite{anderson2018bottom} for image caption and VQA, where the concept of bottom-up and top-down attention is first introduced into the field of computer vision. Inspired by the concept of bottom-up and top-down attention, we design a novel model called K-BAN for the HOI detecion task.

Human-object interaction (HOI) detection aims to detect and recognize how the person in an image interacts with the surrounding objects.
 \begin{figure*}
\begin{center}
 \includegraphics[width=6in]{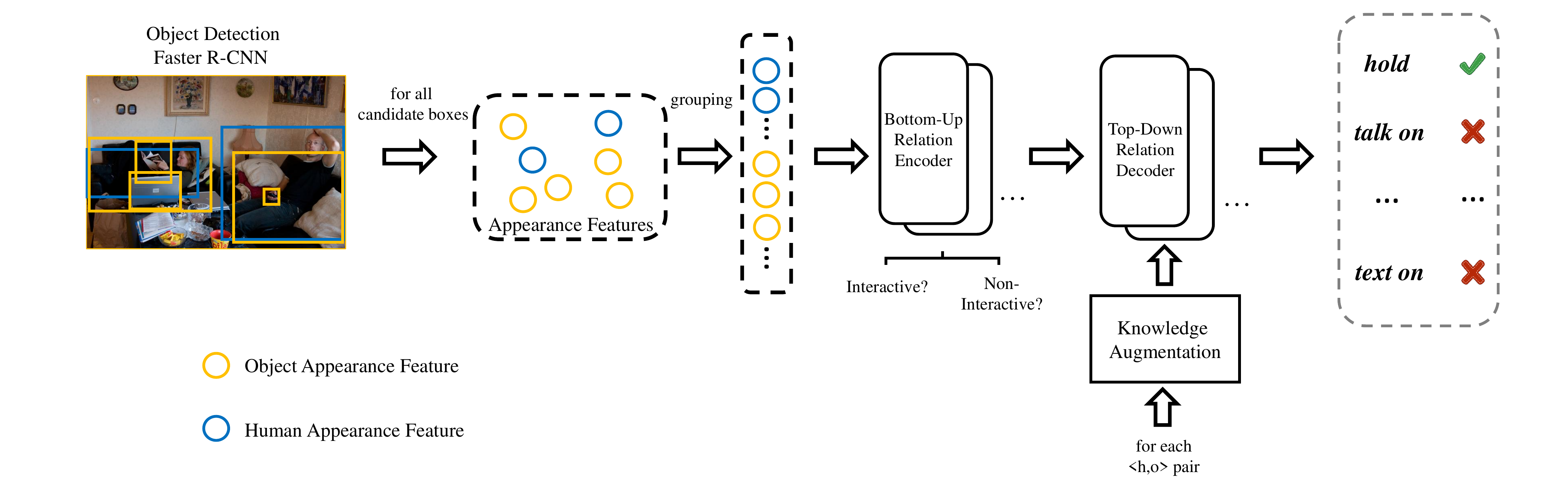}
\end{center}
   \caption{An overview of our method. We utilize a two-stage pipeline for HOI detection. The main body of K-BAN is an encoder-decoder based relation parsing model.}
\label{arch}
 \vspace{-1\baselineskip}
\end{figure*}
 In early works, researchers classify the interaction through the visual appearance information as well as the spatial relationship of a human-object pair\cite{chao2018learning,gupta2015visual,gkioxari2018detecting}. The appearance features of the objects and humans are typically extracted from the backbone of a pre-trained object detector via ROI pooling. Then, the appearance features and the spatial coding of the pair are fed into the models as different streams for the final prediction.

 More recently, the multi-stream frameworks are extended by introducing more diverse information into the HOI detection, such as human poses \cite{fang2018pairwise,wan2019pose}, scene features \cite{liuamplifying,zhou2019relation}, optical flows \cite{liuamplifying}, 3D body shape \cite{li2020detailed} and so on. Fang \etal \cite{fang2018pairwise} introduce the body-part information for the HOI recognition task by capturing the correlation between body-parts. In \cite{wan2019pose}, Wan \etal propose a multi-level relation detection strategy that utilizes human key points to capture global spatial configurations of relations and to extract more fine-grained features at the human part level. In RPNN \cite{zhou2019relation}, the scene features are extracted from the background of images and utilized together with human poses to enhance the representation of humans and objects. Liu \etal \cite{liuamplifying} employ the segmentation masks for the human parts to construct fine-grained layout representation of the object and human. Moreover, they further represent the scene with the word embedding of the scene category estimated by the model pre-trained for scene recognition, and exploit the cues from the optical flows of the image predicted by a pre-trained Im2flow \cite{gao2018im2flow} model.

 Another significant progress in recent years is the advanced exploration on the relation parsing for HOI \cite{fang2018pairwise,gao2018ican,gao2020drg,qi2018learning,zhou2019relation,liu2020consnet}. Fang \etal \cite{fang2018pairwise} utilize the spatial relationship between the human and object as a query to learn the attention on different body-parts of the human. In \cite{gao2018ican}, Gao \etal employ an instance-centric attention module that learns to dynamically highlight regions in an image conditioned on the appearance of each instance, and selectively aggregate features relevant for recognizing HOIs base on the attention. In \cite{gao2020drg}, they further define a dual relation graph, and obtain human-centric and object-centric subgraphs for relational modeling. 
 Given an image, Qi \etal \cite{qi2018learning} propose the GPNN to infer a parse graph that includes the HOI graph structure represented by an adjacency matrix and the node labels. In \cite{zhou2019relation}, Zhou \etal construct ``human-bodypart graph", ``object-bodypart graph" for the relation parsing. In order to incorporate more structural knowledge into HOI, Ye \etal \cite{liu2020consnet} encode the relations among objects, actions and interactions into an undirected graph called consistency graph, and exploits Graph Attention Networks (GATs) to propagate knowledge among HOI categories as well as their constituents.

 However, these multi-stream based works execute the relation parsing in a bottom-up manner independently without the direct guidance of human-object pairs, which fails to fully exploit the power of relation parsing.
To handle the defects, we introduce the knowledge guided top-down attention into HOI, and combine it with the bottom-up attention for relation parsing.

 There are also researchers that have other insights on the HOI detection task. In  \cite{liao2020ppdm,wang2020learning,kim2020uniondet,fang2020dirv}, researchers focus on constructing faster HOI detection models with competitive performance, and propose the one-stage HOI detection methods . Typically, they capture the interaction directly by defining the interaction area, and assign the interaction to the detected instances. Hou \etal \cite{hou2020visual} and Bansal \etal \cite{bansal2020detecting} pay more attention to the solving of long-tail problem in HOI.
 \section{Methodology}
\subsection{Overview} \label{overview}
As shown in Fig.\ref{arch}, our method utilizes a two-stage pipeline for HOI detection, where object detection is executed followed by the pair-wise HOI recognition. Following convention \cite{gao2018ican}, we employ an off-the-shell Faster R-CNN \cite{ren2015faster} pre-trained on COCO dataset \cite{lin2014microsoft} for the detection of human/object instances. We denote the set of detected humans as $\mathbb{H}$, the set of objects as $\mathbb{O}$, and the union of the two set as $S$. Each detected human (object) instance is represented by its bounding box denoted as $b_{h}$ ($b_{o}$) and the confident score $s_{h}$ ($s_{o}$) given by the detector. 
According to the bounding boxes, we extract the appearance features of the detected instances from the backbone module through RoI Pooling. 
For each human, we extract its pose map as in \cite{li2019transferable}, where the estimated human pose is represented as a line-graph. We further construct a spatial map, where the the relative spatial information of the pair is represented by two binary masks of the human and object in their union space. We denote the pose map as $p$ and the spatial map as $sp$.

As shown in Fig.\ref{arch}, the main body of K-BAN is an encoder-decoder based relation parsing model. The encoder serves for the bottom-up scene context modeling. It is constructed upon our newly proposed Group-aware Parsing Modules. The module takes the features of all the detected instances in an image as inputs, and enhances the features through relation parsing. In the module, the interactiveness between each human-object pair is predicted during the relation parsing process so as to introduce the interactiveness knowledge into the relation parsing and operate non-interaction suppression \cite{li2019transferable}. More details about the encoder can be found in Section \ref{sec-en}.

In the decoder, given a pair of $\langle human, object\rangle$ proposal, we combine the semantic features of the object with the spatial coding of the pair as a query. In order to leverage the verb-object co-occurrence knowledge to guide the attention of the model, we further duplicate each query, and augment its duplicates as a set of queries with the verb embeddings of corresponding interactions. These queries are fed into the decoder of the model to guide clue searching on the outputs of the encoder. The clues are then applied to estimate the interaction score $s_{r}$ of the pair. For more details about the decoder, please refer to the Section \ref{sec-de}.

Besides the interaction score $s_{r}$, our model also predicts another interaction score $s_{c}$ from the concatenation of human features, object features and their spatial features of the given pair, which provides a complementary information for relation parsing.
At last, the two scores are fused together with the confidence scores of the instances for the calculation of the overall interaction score $S_{verb}$:
\begin{equation}
        S_{verb} = s_{h} \cdot s_{o} \cdot (s_{r} + s_{c}) / 2
\end{equation}

\begin{figure}[t]
\begin{center}
\includegraphics[width=0.9\linewidth]{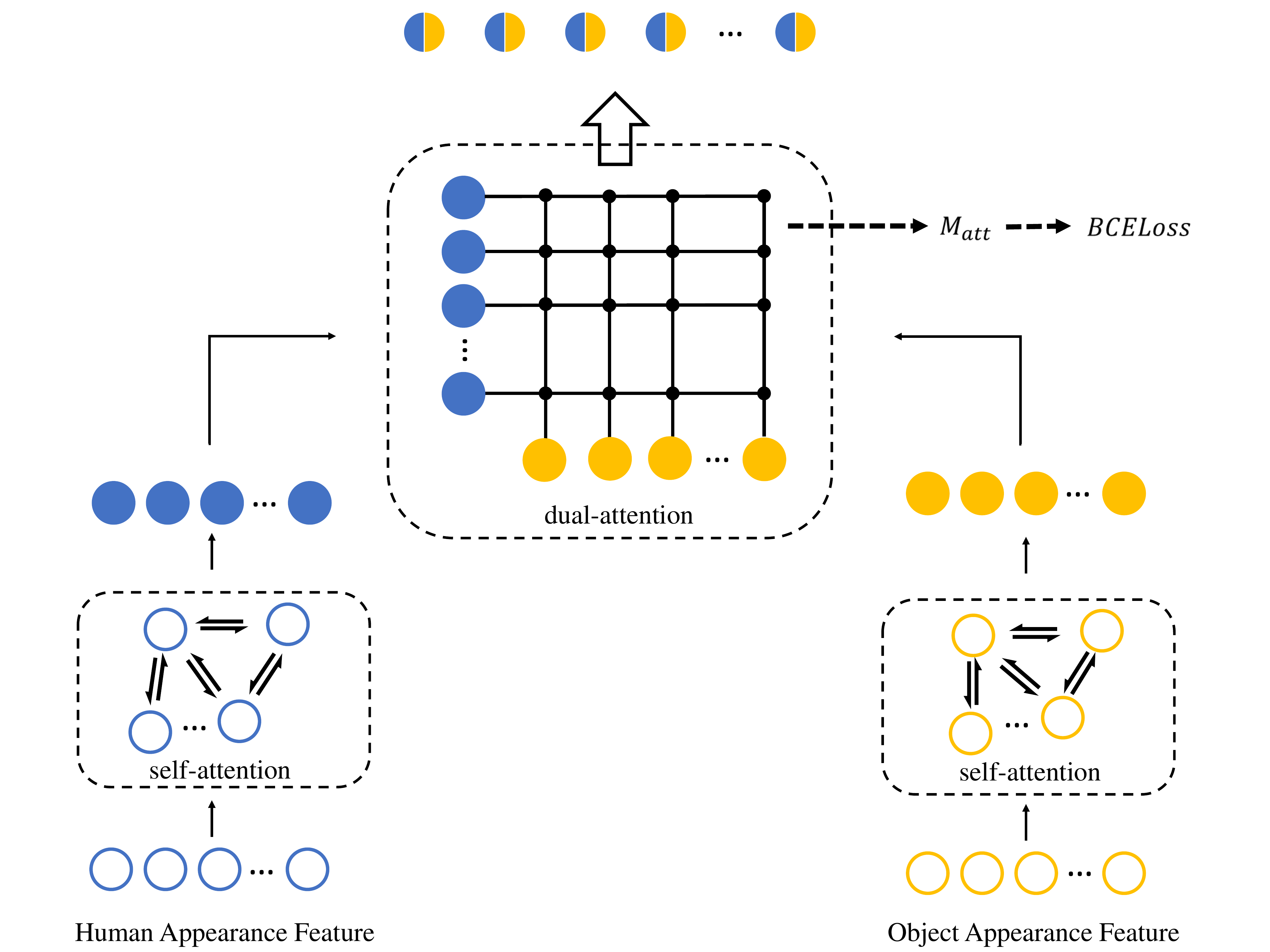}
\end{center}
   \caption{An illustration of the Group-aware Parsing Module. The features of the detected instances are first divided into an object group and a human group. The self-attention is employed for intra-group relation parsing. The dual-attention is for inter-group relation parsing.}
\label{enc}
 \vspace{-1\baselineskip}
\end{figure}
\subsection{Bottom-Up Relation Encoder} \label{sec-en}
The encoder of our model serves for the scene parsing in a bottom-up manner, which can be formulated as $P(A_{i}|\mathbb{H}, \mathbb{O}, i)$, where $i, A_{i}$ denote a detected instance, the relation between the instance and other instances (\ie the attention distribution), respectively. 
An encoder is composed of a stack of identical encoder layers. It takes the features of all the detected instances as inputs, and outputs the features enhanced with scene context information. The input features of an instance are the summation of its appearance features (\ie $v_{o}$ or $v_{h}$) and position coding. Unlike the spatial coding for the spatial stream, the position coding is simply generated by a fc layer that projects a 5-d vector \footnote{We construct the 5-d vector with the normalized top-left coordinates, width, height of the bounding box and the fraction of image area covered by the box.} of a bounding box to a vector in the same dimension as the appearance features.

In order to facilitate the relation parsing with interactiveness knowledge, we propose a Group-aware Parsing Module (GPM) as the encoder layer.
As shown in Fig.\ref{enc}, 
in the module, the features of the detected instances are first divided into an object group and a human group. Two multi-head self-attention layers are adopted for the intra-group relation parsing among the instances in the object group and human group, respectively. Then, the enhanced features of instances in both groups are fed into a multi-head dual-attention layer, where the object attention and the human attention are calculated separately. When calculating the human attention conditioned on objects, we take the object instances as queries $Q_{o}$ and human instances as keys $K_{h}$ and values $V_{h}$, and vise versa for object attention conditioned on humans. In a single head dual-attention layer, given the $Q \in R^{N \times d}, K \in R^{M \times d}, V \in R^{M \times d}$, we compute the outputs as:
\begin{equation}
        Attention(Q,K,V) = max(sigmoid(\frac{QK^{T}}{\sqrt{d}})\odot V)
\end{equation}
where $d$ is the dimension of features and $\frac{1}{\sqrt{d}}$ serves as a scaling factor. $N$ and $M$ represent the number of instances in the two groups. $\odot$ indicates the element-wise multiplication with broadcast
, where the two tensors are expanded to the size of $N \times M \times d$.  The $``max"$ operation is the max pooling that pools the result of the multiplication alone the second dimension. Notice that the attention matrix $Q_{o}K_{h}^{T}$ is the transpose of the matrix $Q_{h}K_{o}^{T}$. Therefore, in practice, we calculate the attention matrix $M_{att}=Q_{o}K_{h}^{T}$ first, and then calculate the dual attentions as:
\begin{equation}
        Attention(Q_{o},K_{h},V_{h}) = max(sigmoid(\frac{M_{att}}{\sqrt{d}}) \odot V_{h})
\end{equation}
\begin{equation}
        Attention(Q_{h},K_{o},V_{o}) = max(sigmoid(\frac{M_{att}^{T}}{\sqrt{d}}) \odot V_{o})
\end{equation}
, respectively. The outputs of dual attention are utilized for the update of $Q$ via a residual connection\cite{he2016deep}.

With a group-aware parsing strategy as mentioned above, we are able to introduce the interactiveness knowledge into the relation parsing by applying the Binary Cross Entropy Loss (BCELoss) on the attention matrix $M_{att}$:
\begin{equation}
        L = BCELoss(M_{att}, GT),
\end{equation}
where $GT \in R^{N \times M}$ is the groundtruth matrix. $GT_{i,j}=1$ when there exists an interaction between the human-object pair, otherwise $GT_{i,j}=0$. In this case, the attention matrix can be also considered as a interactiveness score matrix, where the value of $M_{att}^{i,j}$ is taken as the score of the corresponding human-object pair. We utilize $M_{att}$ to suppress non-interaction pairs in the inference phase.
\begin{figure}[t]
\begin{center}
\includegraphics[width=1\linewidth]{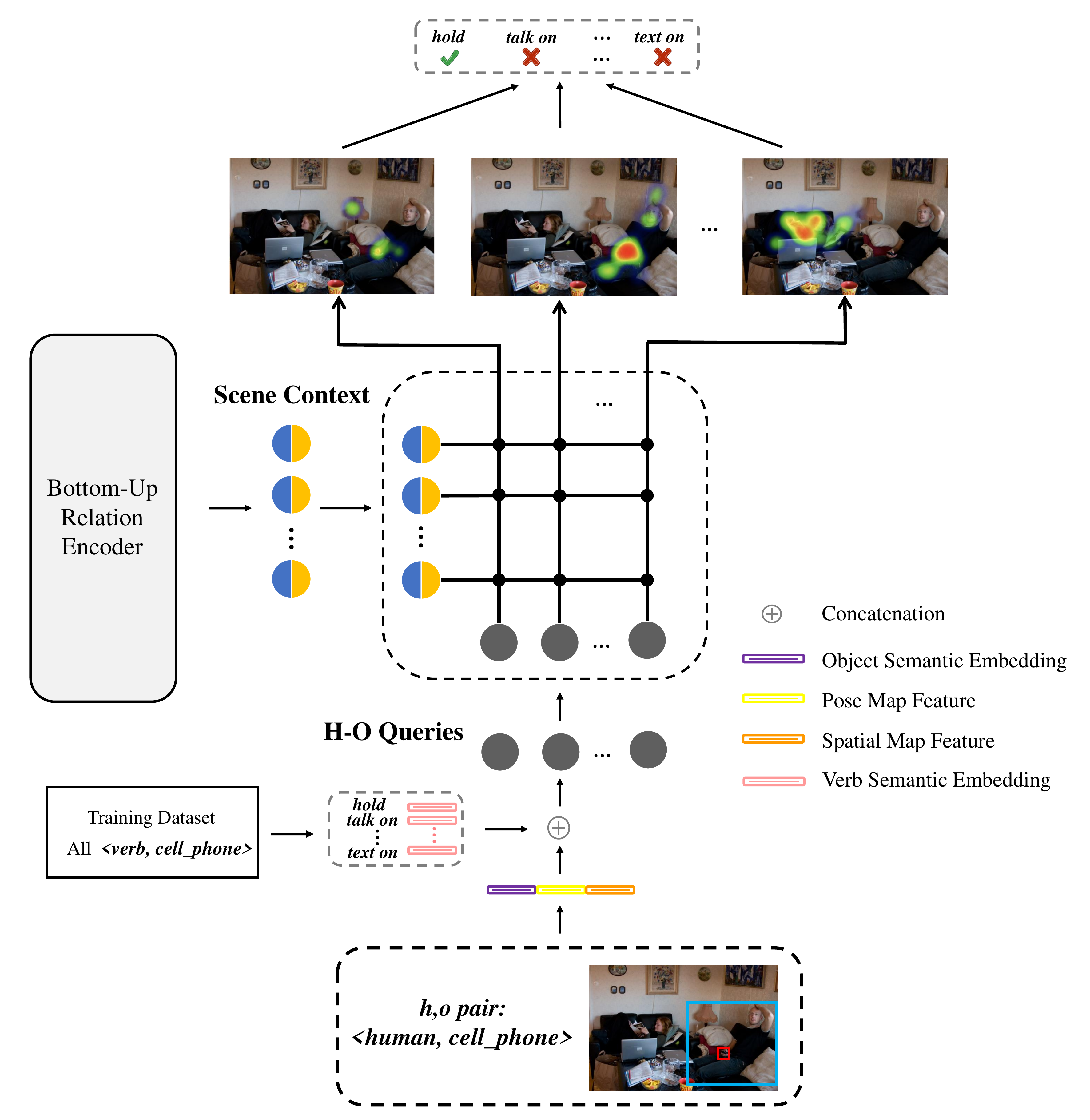}
\end{center}
   \caption{An illustration of the top-down relation decoder.}
\label{dec}
\vspace{-1\baselineskip}
\end{figure}
\subsection{Top-Down Relation Decoder}\label{sec-de}
 Our decoder is composed of a stack of identical decoder layers. Given a $\langle human, object\rangle$ pair, the decoder layers take the pair as a query, and search visual clues for the judgement of the interaction through top-down relation parsing.
We construct the query with the semantic word embedding of the object, pose map $p$ and spatial map $sp$. We utilize Glove \cite{peters2018deep} for the extraction of word embedding. The pose map $p$ and spatial map $sp$ are fed into two convolutional layers for feature learning, and the features are then flattened and concatenated with the word embedding. Moreover, considering that different categories of possible interactions corresponding to the object can be characterized by different visual cues, a preferred decoder should have the capability to adaptively collect different clues for the discrimination of different interactions. Therefore, we further introduce the verb-object co-occurrence as the prior knowledge to guide the attention of the model. Given an object $o$, we denote the verb set, in which the verbs have co-occurred with the object in the training set, as $Verb_{o}$. As shown in Fig.\ref{dec}, we duplicate the query of the given $\langle human, object\rangle$ pair, and concatenate each of its duplicate with the embedding of a verb in $Verb_{o}$. We denote these verb prior knowledge augmented queries as $Q \in R^{N_{verb} \times d_{q}} $, where $N_{verb}$ and $d_{q}$ denote the size of $Verb_{o}$ and the dimension of the query embedding, respectively. Then, the top-down relation parsing can be formulated as $P(A_{\langle h,o\rangle, verb_{o}}|S,\langle h,o\rangle, verb_{o}, p, sp)$. Given the queries $Q$, our decoder takes the instance features enhanced with the scene context by the encoder as keys $K$ and values $V$. The outputs of the decoder are a collection of features (clues) corresponding to different interaction verbs:
 \begin{equation}
        Attention(Q,K,V) = softmax(\frac{QK^{T}}{\sqrt{d}})V
\end{equation}
 The outputs of attention module are utilized for the update of $Q$ via a residual connection\cite{he2016deep}. Inspired by \cite{zhong2020polysemy}, we transform the multi-label verb classification into a set of binary classification problems, where each binary classifier is used for the verification of one specific verb category. The output features of the decoder are fed into the corresponding binary classifiers for verb prediction, which avoids additional post-processing for the fusion of results predicted by each of features.
\subsection{Model Learning}\label{sec-obj}
Given a huamn-object pair $\langle h, o\rangle$, we first construct the binary verb label vector $l$ for the pair, where $l_{i} =  1$ if the $i$th interaction exists between the pair, otherwise $l_{i} = 0$. Given the verb label vector (\ie $l$) and the set of verbs co-occurred with the object (\ie $V_{o}$), the training objective function of our K-BAN is defined as follows:
\begin{align}
    L_{\langle h, v, o\rangle} &= BCELoss(M_{att}^{h,o}, GT_{h,o}) + BCELoss(s_{c},l)\nonumber\\
    &+ \sum_{i \in {V_{o}}}{BCELoss(s_{r}^{i}, l_{i})},
\end{align}
where the first term is the loss function for the binary interactive/non-interactive classification as introduce in Section \ref{sec-en}. The other two terms are the classification losses of the prediction $s_{c}$ and $s_{r}$.
\section{Experimental Results}
\begin{table*}[t]
\centering
\begin{tabular}{c|c|c|c|ccc|ccc}
\hline
\multirow{3}*{Method}&\multirow{3}*{Backbone}&V-COCO&\multirow{3}*{Detector}&\multicolumn{6}{c}{HICO-DET}\\
  &&\multirow{2}*{\makecell[c]{$AP_{role}$\\ Scenario 1}}& & \multicolumn{3}{c|}{Default} & \multicolumn{3}{c}{Known Object}\\
&&&& Full & Rare & \makecell[c]{Non-Rare} & Full & Rare & \makecell[c]{Non-Rare}\\
\hline
InteractNet \cite{gkioxari2018detecting} & R50FPN  & 40.0 & - &-	& -	& -& -	& -	& -\\
iCAN \cite{gao2018ican}& R50 &45.3 & COCO & 14.84 &	10.45	&16.15&	16.26&	11.33&	17.73\\
TIN \cite{li2019transferable}& R50&47.8 &COCO & 17.03&	13.42&	18.11&	19.17&	15.51&	20.26\\
Cascaded \cite{zhou2020cascaded} & R50  & 48.9 & - &-	& -	& -& -	& -	& -\\
Analogies \cite{Peyre_2019_ICCV}   & R50FPN  & - &COCO & 19.40 & 14.60 & 20.90& -	& -	& -\\
PMFNet \cite{wan2019pose}& R50FPN&52.0 &COCO & 17.46	& 15.65& 	18.00& 	20.34& 	17.47	& 21.20\\
DRG \cite{gao2020drg}& R50FPN& 51.0 &COCO & 19.26 &	17.74&	19.71&	23.40&	21.75&	23.89\\
VCL \cite{hou2020visual}& R50& 48.3 &COCO & 19.43& 16.55 & 20.29 &22.00 &19.09 &22.87\\
FCMNet \cite{liuamplifying}& R50 & 53.1 & COCO &20.41&	17.34	&21.56	&22.04	&18.97&	23.12\\
ConsNet\cite{liu2020consnet}& R50FPN &53.2 & COCO &22.15 &17.12&	23.65 & - &	- & -\\
IDN \cite{NEURIPS2020_3493894f} & R50  & 53.3 & COCO & \textbf{23.36}& \textbf{22.47} &\textbf{23.63} &\textbf{26.43} &\textbf{25.01} &\textbf{26.85}\\
\textbf{K-BAN}& R50& \textbf{53.50} &COCO & 20.01 & 14.25 &21.73& 23.08 & 16.57&25.03\\
\hline
GPNN \cite{qi2018learning} & R101 &44.0 & COCO & 13.11& 	9.34	& 14.23	& -& 	-& 	-\\
No-Frills \cite{gupta2019no}& R152& - & COCO& 17.18& 	12.17& 	18.68	& -	& -	& -\\
VSGNet \cite{Ulutan_2020_CVPR}&R152 & 51.76& COCO & 19.80&	16.05&	20.91	&-	&-&	-\\
ACP \cite{kim2020detecting}& R152 &52.98 & COCO &20.59&	15.92&	21.98&	-&	-&	-\\
\textbf{K-BAN}& R152& \textbf{53.70}&COCO & \textbf{21.48} & \textbf{16.85} &\textbf{22.86}& \textbf{24.29} & \textbf{19.09} & \textbf{25.85}\\
\hline
\hline
UniDet$ ^\dag$ \cite{kim2020uniondet} & R50FPN  & 47.5 &  HICO & 17.58 & 11.72 & 19.33 & 19.76 & 14.68 & 21.27\\
TIN \cite{li2019transferable}& R50&- &HICO & 23.17 &15.02& 25.61 &24.76 &16.01 &27.37\\
VCL \cite{hou2020visual}& R50& - &HICO & 23.63& 17.21& 25.55 &25.98 &19.12 &28.03\\
DRG \cite{gao2020drg}& R50FPN& - &HICO & 24.53& 19.47& 26.04 &27.98 &23.11 &29.43\\
IDN \cite{NEURIPS2020_3493894f} & R50  & - & HICO & 26.29 &\textbf{22.61} &27.39 &28.24 &\textbf{24.47} &29.37\\
\textbf{K-BAN}& R50& - &HICO & \textbf{27.01} & 18.14 &\textbf{29.66}& \textbf{29.54} & 19.74 & \textbf{32.47}\\
\hline
Wang \etal $ ^\dag$ \cite{wang2020learning}  & H104  & 51.0 & HICO & 19.56& 12.79& 21.58 & 22.05	& 15.77	& 23.92\\
PPDM$ ^\dag$ \cite{liao2020ppdm}  & H104  & - & HICO & 21.10	& 14.46	& 23.09 & -	& -	& -\\
DIRV $ ^\dag$ \cite{fang2020dirv} & EfficientDet  & \textbf{56.1} & HICO & 21.78& 16.38& 23.39& 25.52& 20.84& 26.92\\
\textbf{K-BAN}& R152& -&HICO & \textbf{28.83} & \textbf{20.29} & \textbf{31.31} & \textbf{31.05} & \textbf{21.41} & \textbf{33.93}\\
\hline
\hline
iCAN \cite{gao2018ican}& R50 &- & GT & 	33.38	&21.43	&36.95& -	& -	& -\\
Analogies \cite{Peyre_2019_ICCV}    & R50FPN  & - &GT& 34.35 & 27.57 & 36.38 & -	& -	& -\\
IDN \cite{NEURIPS2020_3493894f} & R50  & - & GT & 43.98 &\textbf{40.27}& 45.09 &- &- &-\\
\textbf{K-BAN}& R50& - &GT & \textbf{50.38} & 30.16 & \textbf{56.42}& -	& -	& -\\
\textbf{K-BAN}& R152& - &GT & \textbf{52.99} & 34.91 & \textbf{58.40} & -	& -	& -\\
\end{tabular}
\caption{Results comparison on the test sets of HICO-DET and V-COCO dataset. $ ^\dag$ indicates one-stage models.}
\label{hico}
 \vspace{-1\baselineskip}
\end{table*}
\subsection{Experimental Setup}
\textbf{Datasets.} We evaluate our method on the V-COCO \cite{gupta2015visual} and HICO-DET \cite{chao2018learning} datasets. The V-COCO (Verbs in COCO) dataset, which contains a total of 10,346 images annotated with 16,199 person instances, are divided into three splits(2,533 in train set, 2,867 in validation set and 4,946 in test set). Binary labels for 29 different action classes (five of them have no interactive object, \eg ``stand", ``walk") are assigned to each person instance. HICO-DET (Humans Interacting with Common Objects) dataset consists of 47, 774 images (38,118 for training and 9658 for testing), including 600 human-object interaction categories over 117 common actions (including ``no interaction" class) performed on 80 objects introduced in the MS-COCO dataset. Note that due to the insufficient number of training samples (less than 10), among all human object interactions, 138 interactions are attributed to rare categories. Correspondingly, the remaining 462 interactions are non-rare, and all 600 interactions constitute a complete category. We report the performance of these three categories following previous work.

\textbf{Evaluation Metrics.} For both V-COCO and HICO-DET datasets, we use the mean Average Precision (mAP) to measure the performance of methods. A triplet $\langle human, verb, object\rangle$ is correct if and only if 1) the overlap between detected bounding boxes (for both human and object) and the corresponding ground truths are greater than 0.5 and 2) the class of the verb is correctly predicted.

\textbf{Implementation Details.} For a fair comparison, we use Faster-RCNN \cite{ren2015faster} pretrained on COCO dataset to detect human and object instances in images. We adopt the backbone of the detector as our feature extractor. During training, a SGD optimizer is used to update the model parameters. For V-COCO, we train the network for 310k iterations with an initial learning rate of 1e-3 and a weight decay of 5e-4. For HICO-DET, we set 5e-2 as the initial learning rate and reduce it to 5e-3 at 800k iteration, then reduce it to 5e-4 at 950k iteration, and finally stop training at 1200k. In the inference stage, we only pair the detected human and object boxes that have detection scores higher than threshold $t$ ($t = 0.4$ for human and $0.1$ for object). For HICO-DET, the thresholds for human and object are set to 0.6 and 0.1, respectively. The scores of the attention matrix $M_{att}$ between groups are used to filter non-interactive pairs.
\subsection{Comparisons with State-of-the-Art Methods}\label{sota}
In this section, we compare our method with current state-of-the-art methods on V-COCO and HICO-DET datasets. 
As shown in Table.\ref{hico}, on V-COCO dataset, our model(Resnet50) achieves 53.50 mAP, which surpass all the other competitors that employ Resnet50 or the stronger ones (\eg Resnet50-FPN/Resnet152) as their backbones. 
 Current state-of-the-art method on V-COCO is DIRV \cite{fang2020dirv}, which is an one-stage method. Unlike the two-stage methods that utilize COCO pretrained detector to generate human-object pair proposals directly, one-stage methods \eg DIRV finetune the detector on V-COCO and don't require generating pair-wise proposals. With a stronger backbone \ie EfficientDet, it achieves a SOTA result of 56.1 mAP.\\
HICO-DET is a large scale dataset. Finetuning the object detector on the dataset can improve the result significantly. Therefore, on the HICO-DET dataset, we report the results achieved under 3 different settings to evaluate the models fairly and comprehensively. In the first setting, all the methods utilize a COCO pretrained detector, and operate interaction detection based on the detection results of the detector. In the second setting, the detector is further finetuned on the HICO-DET dataset to provide more accurate detection results. The one-stage methods that train/finetune the object detector on HICO-DET are also classified as the second setting. In the third setting, the groud-truth pairs are utilized for interaction classification, thereby eliminating the influence of the object detector and focusing on the evaluation of the interaction discrimination ability of the models.

 Under the first setting, our method achieves a comparable performance among the methods that utilize Resnet50 as the backbone, and has a competitive results compared to the methods implemented with Resnet152. IDN\cite{NEURIPS2020_3493894f} achieves the state-of-the-art methods among all the methods under the first setting. Under the second setting, our method outperforms IDN and achieves the SOTA performance. It shows that in the situation that has high quality detection results, our method has stronger representation learning capability for the discrimination of human-object interactions. The conclusion reveals that with the advancement of object detector, our method would show greater potential in HOI. Moreover, eliminating the influence of the object detector by taking the groudtruth pairs as inputs, our method outperforms IDN by a large margin under the third setting. The results further demonstrate the conclusion as well as the superiority of our method. We also notice that IDN consistently outperforms our method on the Rare subset of HICO-DET. It reveals that IDN is more robust to the long-tail problem in HOI. It will be our next work to improve the performance of our method on the recognition of tail classes.

  For the runtime performance, the inference speed of K-BAN reaches 7.49 FPS on HICO-DET. As a comparison, speeds of other methods are as follows: No-Frills(2.02 FPS), PMFNet(3.95 FPS), iCAN(4.90 FPS), DRG(5.0 FPS), IDN(10.04 FPS) and PPDM(14.08 FPS). In general, one-stage methods \eg PPDM have higher runtime-efficiency. Our model achieves competitive performance in terms of speed and accuracy on the public benchmarks.
 \begin{table}
\centering
\begin{tabular}{cccc|c}
\hline
Enc  & \makecell[c]{GPM} & Dec & \makecell[c]{KA} & $AP_{role}$\\
\hline
$\checkmark$& $\checkmark$ & - & -& 52.42 \\\hline
-& - & $\checkmark$ & $\checkmark$ & 53.43 \\\hline
$\checkmark$ & - & $\checkmark$ & $\checkmark$ & 53.68 \\\hline
$\checkmark$ & $\checkmark$  & $\checkmark$ & - & 53.84 \\\hline
$\checkmark$ & $\checkmark$  & $\checkmark$ & $\checkmark$ & \textbf{54.28} \\
\end{tabular}
\caption{Ablation study on the validation set of V-COCO.}
\label{ablation}
\end{table}
\begin{table}
\centering
\begin{tabular}{c|c|c}
\hline
Arch&$\sharp$Enc + $\sharp$Dec & $AP_{role}$\\
\hline
\multirow{2}*{B-U Only}&$2Enc$ & 52.42 \\
&$3Enc$ & 52.37\\
\hline
\multirow{2}*{T-D Only}& $2Dec$ & 53.43\\
& $3Dec$ & 53.52\\
\hline
\multirow{4}*{B-U + T-D}&$1Enc+1Dec$ & 54.25\\
&$1Enc+2Dec$ & 54.23\\
&$2Enc+1Dec$ & 53.28\\
&$2Enc+2Dec$ & \textbf{54.28}\\
\end{tabular}
\caption{Comparisons of models with different architectures. ``B-U" indicates Bottom-Up attention. ``T-D" denotes Top-Down Attention. Results are reported on the validation set of V-COCO.}
\label{ablation2}
 \vspace{-1\baselineskip}
\end{table}

 \begin{figure*}[t]
\begin{center}
 \includegraphics[width=7in]{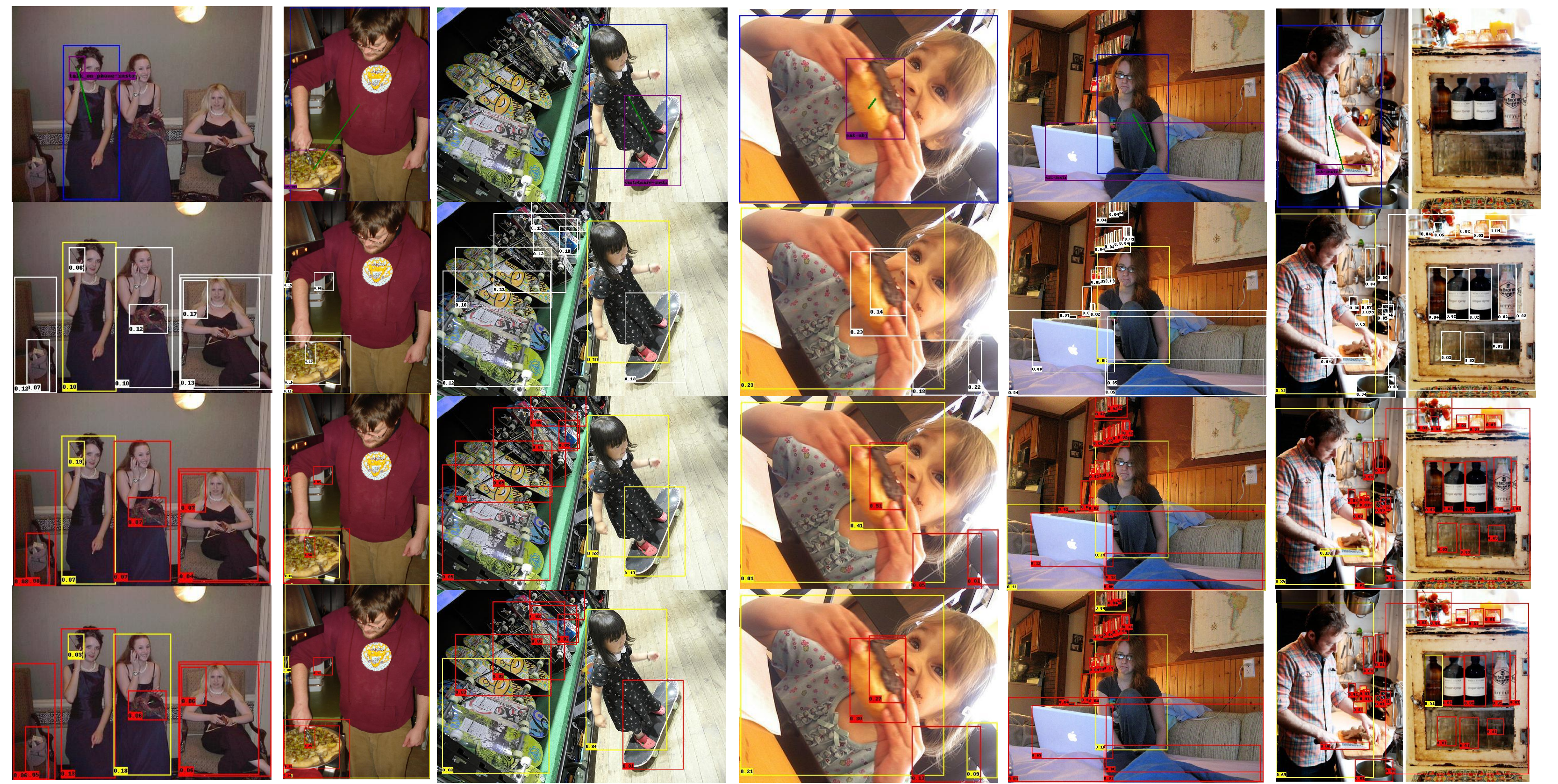}
\end{center}
   \caption{Qualitative visualization results of our method. We present the results on 6 different samples in this figure, where each column corresponds to the results of one sample.}
\label{vis}
 \vspace{-1\baselineskip}
\end{figure*}

\subsection{Ablation Study}\label{ab_sec}
In this section, we conduct ablation study to justify the contributions of different modules in this work, including the bottom-up relation encoder (Enc), Group-aware Parsing Module (GPM), top-down relation decoder (Dec) and knowledge augmentation for decoder queries (KA). We utilize V-COCO for the ablation study, where the models are trained on the train set and tested on the validation set of the V-COCO. As shown in Table.\ref{ablation}, with the encoder implemented with Group-aware Parsing Module, our model achieves 52.42 mAP on the validation set. With only the decoder that takes knowledge augmented queries and the original instance features from the backbone as inputs, the model outperforms the encoder-only model by 1.01 mAP. Combining our bottom-up relation encoder and top-down relation decoder, our full model achieves 54.28 mAP. Removing the Group-aware Parsing Module in the encoder and utilizing the self-attention mechanism for relation parsing of detected instances without grouping, the performance of our model drops by 0.6, which demonstrates the effectiveness of the Group-aware Parsing Module. We also witness a performance drop of 0.44 when we remove the knowledge augmentation for the human-object query. It reveals that verb semantic information is beneficial to the top-down clue mining.

For better understanding our model, we further construct our model with different architectures. As shown in Table.\ref{ablation2}, we vary the number of encoder/decoder layers, and test their performance on the validation set of V-COCO. We can see that the result of $2Dec$ is better than the result of $2Enc$, which is consistent with the result in the ablation study. It reveals that the top-down attention is more helpful than the bottom-up attention in the relation parsing for HOI. Additionally, simply increasing the encoder layers or decoder layers (\ie $3Enc$ and $3Dec$), the performance improvement is trivial. Moreover, when we employ the architecture of $1Enc+1Dec$, the total parameter number of the model is comparable with $2Enc$ and $2Dec$, but the performance of the model is improved to 54.25mAP. It demonstrates that the bottom-up attention and top-down attention in our model is complementary. Moreover, we can see that our model is robust to the choice of different compositions of the number of encoders and decoders.


\subsection{Qualitative Evaluation}\label{QE}
In this section, we present the qualitative visualization results for the better understanding of our work. We try to answer the following questions through the visualization results: 1) what is the difference between the distribution of top-down attention and bottom-up attention? 2) given a non-interactive human-object pair as query, what dose the top-down attention distribution of the model look like?

We present the results on 6 different samples in Fig.\ref{vis}, where each column corresponds to the results of one sample. For each sample, one of the goundtruth HOI triplets in the image is shown in the first row. The second row illustrates the visualization results for the bottom-up attention learnt by the encoder module with self-attention (\ie a model w/ Enc and w/o GPM in the ablation study). In the third row, we show the correctly detected HOI triplets as well as the top-down attention of our model. The forth row shows the non-interactive pairs suppressed by our method and the top-down attention distribution conditioned on the pairs. From row 2 to row 4, the queries are indicated by yellow boxes.

As shown in Fig.\ref{vis}, the top-down attention tends to focus on the most related instances while the bottom-up attention is distributed more equally among all the instances in scenes. It shows that top-down attention can better conduct the pair-specific relation parsing to filter the redundant or less relevant information, while the bottom-up attention gives a more comprehensive modeling of the entire scene. We also notice that, given a non-interactive pair, the decoder tends to focus on the human itself rather than other object instances. For example, in the first column, our decoder pays more attention on the phone which is given in the query with the women using it (row 3), and the attention is distracted from the phone when it is paired as a query with another woman (row 4). It reveals that the top-down attention may reduce the misleading from non-interactive objects in scenes. 
\section{Conclusions}
In this paper, we propose a new model called K-BAN for the HOI detection. The model unify both the bottom-up attention and knowledge-guided top-down attention in a single encoder-decoder architecture to improve the relation parsing and interaction representation learning in HOI. Given the detected instances in a scene, the encoder serves for the scene parsing in a bottom-up manner. With the guidance of the verb-object co-occurrence knowledge, the decoder collects the clues for the judgement of the interaction between a given human-object pair through top-down relation parsing. We conduct extensive experiments on the V-COCO and HICO-DET datasets, and the experimental results demonstrate the effectiveness of our method.

{\small
\bibliographystyle{ieee_fullname}
\bibliography{egbib}

\begin{thebibliography}{10}\itemsep=-1pt

\bibitem{anderson2018bottom}
Peter Anderson, Xiaodong He, Chris Buehler, Damien Teney, Mark Johnson, Stephen
  Gould, and Lei Zhang.
\newblock Bottom-up and top-down attention for image captioning and visual
  question answering.
\newblock In {\em Proceedings of the IEEE conference on computer vision and
  pattern recognition}, pages 6077--6086, 2018.

\bibitem{bansal2020detecting}
Ankan Bansal, Sai~Saketh Rambhatla, Abhinav Shrivastava, and Rama Chellappa.
\newblock Detecting human-object interactions via functional generalization.
\newblock In {\em Proceedings of the AAAI Conference on Artificial
  Intelligence}, volume~34, pages 10460--10469, 2020.

\bibitem{buschman2007top}
Timothy~J Buschman and Earl~K Miller.
\newblock Top-down versus bottom-up control of attention in the prefrontal and
  posterior parietal cortices.
\newblock {\em science}, 315(5820):1860--1862, 2007.

\bibitem{chao2018learning}
Yu-Wei Chao, Yunfan Liu, Xieyang Liu, Huayi Zeng, and Jia Deng.
\newblock Learning to detect human-object interactions.
\newblock In {\em 2018 ieee winter conference on applications of computer
  vision (wacv)}, pages 381--389. IEEE, 2018.

\bibitem{corbetta2002control}
Maurizio Corbetta and Gordon~L Shulman.
\newblock Control of goal-directed and stimulus-driven attention in the brain.
\newblock {\em Nature reviews neuroscience}, 3(3):201--215, 2002.

\bibitem{fang2018pairwise}
Hao-Shu Fang, Jinkun Cao, Yu-Wing Tai, and Cewu Lu.
\newblock Pairwise body-part attention for recognizing human-object
  interactions.
\newblock In {\em Proceedings of the European Conference on Computer Vision
  (ECCV)}, pages 51--67, 2018.

\bibitem{fang2020dirv}
Hao-Shu Fang, Yichen Xie, Dian Shao, and Cewu Lu.
\newblock Dirv: Dense interaction region voting for end-to-end human-object
  interaction detection.
\newblock In {\em The AAAI Conference on Artificial Intelligence (AAAI)}, 2021.

\bibitem{gao2020drg}
Chen Gao, Jiarui Xu, Yuliang Zou, and Jia-Bin Huang.
\newblock Drg: Dual relation graph for human-object interaction detection.
\newblock In {\em European Conference on Computer Vision}, pages 696--712.
  Springer, 2020.

\bibitem{gao2018ican}
Chen Gao, Yuliang Zou, and Jia-Bin Huang.
\newblock ican: Instance-centric attention network for human-object interaction
  detection.
\newblock In {\em British Machine Vision Conference}, 2018.

\bibitem{gao2018im2flow}
Ruohan Gao, Bo Xiong, and Kristen Grauman.
\newblock Im2flow: Motion hallucination from static images for action
  recognition.
\newblock In {\em Proceedings of the IEEE Conference on Computer Vision and
  Pattern Recognition}, pages 5937--5947, 2018.

\bibitem{gkioxari2018detecting}
Georgia Gkioxari, Ross Girshick, Piotr Doll{\'a}r, and Kaiming He.
\newblock Detecting and recognizing human-object interactions.
\newblock In {\em Proceedings of the IEEE Conference on Computer Vision and
  Pattern Recognition}, pages 8359--8367, 2018.

\bibitem{gupta2015visual}
Saurabh Gupta and Jitendra Malik.
\newblock Visual semantic role labeling.
\newblock {\em arXiv preprint arXiv:1505.04474}, 2015.

\bibitem{gupta2019no}
Tanmay Gupta, Alexander Schwing, and Derek Hoiem.
\newblock No-frills human-object interaction detection: Factorization, layout
  encodings, and training techniques.
\newblock In {\em Proceedings of the IEEE International Conference on Computer
  Vision}, pages 9677--9685, 2019.

\bibitem{he2016deep}
Kaiming He, Xiangyu Zhang, Shaoqing Ren, and Jian Sun.
\newblock Deep residual learning for image recognition.
\newblock In {\em Proceedings of the IEEE conference on computer vision and
  pattern recognition}, pages 770--778, 2016.

\bibitem{hou2020visual}
Zhi Hou, Xiaojiang Peng, Yu Qiao, and Dacheng Tao.
\newblock Visual compositional learning for human-object interaction detection.
\newblock In {\em European Conference on Computer Vision}, pages 584--600.
  Springer, 2020.

\bibitem{kim2020uniondet}
Bumsoo Kim, Taeho Choi, Jaewoo Kang, and Hyunwoo~J Kim.
\newblock Uniondet: Union-level detector towards real-time human-object
  interaction detection.
\newblock In {\em European Conference on Computer Vision}, pages 498--514.
  Springer, 2020.

\bibitem{kim2020detecting}
Dong-Jin Kim, Xiao Sun, Jinsoo Choi, Stephen Lin, and In~So Kweon.
\newblock Detecting human-object interactions with action co-occurrence priors.
\newblock {\em arXiv preprint arXiv:2007.08728}, 2020.

\bibitem{li2020detailed}
Yong-Lu Li, Xinpeng Liu, Han Lu, Shiyi Wang, Junqi Liu, Jiefeng Li, and Cewu
  Lu.
\newblock Detailed 2d-3d joint representation for human-object interaction.
\newblock In {\em Proceedings of the IEEE/CVF Conference on Computer Vision and
  Pattern Recognition}, pages 10166--10175, 2020.

\bibitem{NEURIPS2020_3493894f}
Yong-Lu Li, Xinpeng Liu, Xiaoqian Wu, Yizhuo Li, and Cewu Lu.
\newblock Hoi analysis: Integrating and decomposing human-object interaction.
\newblock In H. Larochelle, M. Ranzato, R. Hadsell, M.~F. Balcan, and H. Lin,
  editors, {\em Advances in Neural Information Processing Systems}, volume~33,
  pages 5011--5022. Curran Associates, Inc., 2020.

\bibitem{li2019transferable}
Yong-Lu Li, Siyuan Zhou, Xijie Huang, Liang Xu, Ze Ma, Hao-Shu Fang, Yanfeng
  Wang, and Cewu Lu.
\newblock Transferable interactiveness knowledge for human-object interaction
  detection.
\newblock In {\em Proceedings of the IEEE Conference on Computer Vision and
  Pattern Recognition}, pages 3585--3594, 2019.

\bibitem{liao2020ppdm}
Yue Liao, Si Liu, Fei Wang, Yanjie Chen, Chen Qian, and Jiashi Feng.
\newblock Ppdm: Parallel point detection and matching for real-time
  human-object interaction detection.
\newblock In {\em Proceedings of the IEEE/CVF Conference on Computer Vision and
  Pattern Recognition}, pages 482--490, 2020.

\bibitem{lin2014microsoft}
Tsung-Yi Lin, Michael Maire, Serge Belongie, James Hays, Pietro Perona, Deva
  Ramanan, Piotr Doll{\'a}r, and C~Lawrence Zitnick.
\newblock Microsoft coco: Common objects in context.
\newblock In {\em European conference on computer vision}, pages 740--755.
  Springer, 2014.

\bibitem{liuamplifying}
Y Liu, Q Chen, and A Zisserman.
\newblock Amplifying key cues for human-object-interaction detection.
\newblock {\em Lecture Notes in Computer Science}.

\bibitem{liu2020consnet}
Ye Liu, Junsong Yuan, and Chang~Wen Chen.
\newblock Consnet: Learning consistency graph for zero-shot human-object
  interaction detection.
\newblock In {\em Proceedings of the 28th ACM International Conference on
  Multimedia}, pages 4235--4243, 2020.

\bibitem{peters2018deep}
Matthew~E Peters, Mark Neumann, Mohit Iyyer, Matt Gardner, Christopher Clark,
  Kenton Lee, and Luke Zettlemoyer.
\newblock Deep contextualized word representations.
\newblock {\em arXiv preprint arXiv:1802.05365}, 2018.

\bibitem{Peyre_2019_ICCV}
Julia Peyre, Ivan Laptev, Cordelia Schmid, and Josef Sivic.
\newblock Detecting unseen visual relations using analogies.
\newblock In {\em Proceedings of the IEEE/CVF International Conference on
  Computer Vision (ICCV)}, October 2019.

\bibitem{qi2018learning}
Siyuan Qi, Wenguan Wang, Baoxiong Jia, Jianbing Shen, and Song-Chun Zhu.
\newblock Learning human-object interactions by graph parsing neural networks.
\newblock In {\em Proceedings of the European Conference on Computer Vision
  (ECCV)}, pages 401--417, 2018.

\bibitem{ren2015faster}
Shaoqing Ren, Kaiming He, Ross Girshick, and Jian Sun.
\newblock Faster r-cnn: Towards real-time object detection with region proposal
  networks.
\newblock In {\em Advances in neural information processing systems}, pages
  91--99, 2015.

\bibitem{Ulutan_2020_CVPR}
Oytun Ulutan, A~S~M Iftekhar, and B.~S. Manjunath.
\newblock Vsgnet: Spatial attention network for detecting human object
  interactions using graph convolutions.
\newblock In {\em IEEE/CVF Conference on Computer Vision and Pattern
  Recognition (CVPR)}, June 2020.

\bibitem{vaswani2017attention}
Ashish Vaswani, Noam Shazeer, Niki Parmar, Jakob Uszkoreit, Llion Jones,
  Aidan~N Gomez, {\L}ukasz Kaiser, and Illia Polosukhin.
\newblock Attention is all you need.
\newblock In {\em Advances in neural information processing systems}, pages
  5998--6008, 2017.

\bibitem{wan2019pose}
Bo Wan, Desen Zhou, Yongfei Liu, Rongjie Li, and Xuming He.
\newblock Pose-aware multi-level feature network for human object interaction
  detection.
\newblock In {\em Proceedings of the IEEE International Conference on Computer
  Vision}, pages 9469--9478, 2019.

\bibitem{wang2020learning}
Tiancai Wang, Tong Yang, Martin Danelljan, Fahad~Shahbaz Khan, Xiangyu Zhang,
  and Jian Sun.
\newblock Learning human-object interaction detection using interaction points.
\newblock In {\em Proceedings of the IEEE/CVF Conference on Computer Vision and
  Pattern Recognition}, pages 4116--4125, 2020.

\bibitem{zhong2020polysemy}
Xubin Zhong, Changxing Ding, Xian Qu, and Dacheng Tao.
\newblock Polysemy deciphering network for human-object interaction detection.
\newblock In {\em Proc. Eur. Conf. Comput. Vis}, 2020.

\bibitem{zhou2019relation}
Penghao Zhou and Mingmin Chi.
\newblock Relation parsing neural network for human-object interaction
  detection.
\newblock In {\em Proceedings of the IEEE International Conference on Computer
  Vision}, pages 843--851, 2019.

\bibitem{zhou2020cascaded}
Tianfei Zhou, Wenguan Wang, Siyuan Qi, Haibin Ling, and Jianbing Shen.
\newblock Cascaded human-object interaction recognition.
\newblock pages 4263--4272, 2020.

\end{thebibliography}
}

\end{document}